\documentclass{article}
\usepackage{spconf,amsmath,amsfonts,graphicx,multirow}

\title{Deep Speaker Embedding Learning with Multi-Level Pooling for Text-Independent Speaker Verification}

\name{Yun Tang, Guohong Ding, Jing Huang, Xiaodong He, Bowen Zhou}
\address{JD AI Research \\
         675 East Middlefield Road \\
         Mountain View, CA 94043, USA}
\begin{document}
%
\maketitle
\begin{abstract}
This paper aims to improve the widely used deep speaker embedding x-vector model. We propose the following improvements: (1) a hybrid neural network structure using both time delay neural network (TDNN) and long short-term memory neural networks (LSTM) to generate complementary speaker information at different levels; (2) a multi-level pooling strategy to collect speaker information from both TDNN and LSTM layers; (3) a regularization scheme on the speaker embedding extraction layer to make the extracted embeddings suitable for the following fusion step. The synergy of these improvements are shown on the NIST SRE 2016 eval test (with a 19\% EER reduction) and SRE 2018 dev test (with a 9\% EER reduction), as well as more than 10\% DCF scores reduction on these two test sets over the x-vector baseline. 
\end{abstract}
\section{Introduction}
\label{sec:intro}
Speaker verification (SV)\cite{hansen2015speaker} is one of the key components for human machine interface and initially was widely used in person identification for security purposes. Nowadays with intelligent speech assistants such as Alexa, Google Home, Siri and Cortana being used in home environments as well as on smartphones, the demand for SV technology is rising. This is especially true for robust speaker verification in challenging acoustic conditions and different population of speakers. 
The speaker verification problem usually falls into two categories: text-dependent (TD) SV and text-independent (TI) SV. In the TD SV system, the transcriptions for the test utterances and enrollment utterances are the same and usually limited to a small set. In the TI SV system, there is no constraint on transcriptions for both test and enrollment utterances. Hence the TI case is more difficult than the TD case due to larger variations introduced by different utterance transcriptions and duration. In this study, we will focus on the more challenging TI speaker verification system.


Recently, more attention has been drawn to the use of deep neural networks (DNN) to generate speaker embedding representations  \cite{heigold2016end} \cite{snyder2017deep} \cite{snyder2018x}. These deep speaker embedding systems are shown to have large improvements over the i-vector \cite{Dehak2011FrontEndFA} based methods, especially the recently proposed x-vector system with data augmentation \cite{snyder2018x}.
The DNN based speaker embedding extraction system usually consists of three components: frame level feature processing, utterance (speaker) level feature processing and training loss.
Frame level processing deals with local short span of acoustic features. It can be done via recurrent neural networks \cite{heigold2016end} or convolutional neural networks \cite{snyder2017deep}\cite{li2018deep}. 
Utterance level processing forms speaker representation based on the frame level output. A pooling layer is used to gather frame level information to form utterance level representation. Methods such as statistical pooling \cite{snyder2017deep}, max pooling\cite{novoselov2018deep}, attentive statistical pooling \cite{okabe2018attentive}, multi-headed attentive statistical pooling \cite{zhu2018self} are popular choices. 
Cross entropy and triplet loss are two widely used training losses. Cross entropy based methods are focused on reducing the confusion for all speakers in the training data \cite{snyder2017deep}, while triplet loss based methods \cite{li2017deep}\cite{Wan2018GeneralizedEL}\cite{novoselov2018triplet}\cite{huang2018angular} are focused on increasing the margin between similar speakers.

In this paper, we propose a novel deep speaker embedding learning framework to improve upon the x-vector system. We first add LSTM layers to the x-vector's TDNN structure, because sequential modeling of an utterance would generate different speaker information from that of TDNN. In order to aggregate these different information, we propose a multi-level pooling
strategy to fuse different information from the different frame level models, one from TDNN and one from LSTM. We also add a regularization term on the speaker embedding extraction layer in order to make the system output more suitable for the back-end processing. Overall, the proposed new speaker embedding system improves the equal error rate (EER) by 19\% and the detection cost function(DCF)\cite{hansen2015speaker} score by 12\%, compared to the previous x-vector baseline on the NIST SRE 2016 eval test. This is to our knowledge the lowest EERs (9.2\% on Tagalog and 3.1\% on Cantonese) on SRE 2016 eval test in publications so far.    

The rest of this paper is organized as follows. Section \ref{sec:relatedwork} briefly describes prior work, including the baseline x-vector system and related work.  The proposed new system is introduced in Section \ref{sec:sys_des} on gathering speaker information from different modeling levels. The experimental set up, results and analysis are described in Section \ref{sec:expt}. Finally, the conclusion is given at Section \ref{sec:con_dis}.
\section{Prior work}\label{sec:relatedwork}
\subsection{Baseline x-vector System}
\label{ssec:baseline}

\begin{figure}[htb]
  \centering
  \centerline{\includegraphics[width=5.5cm]{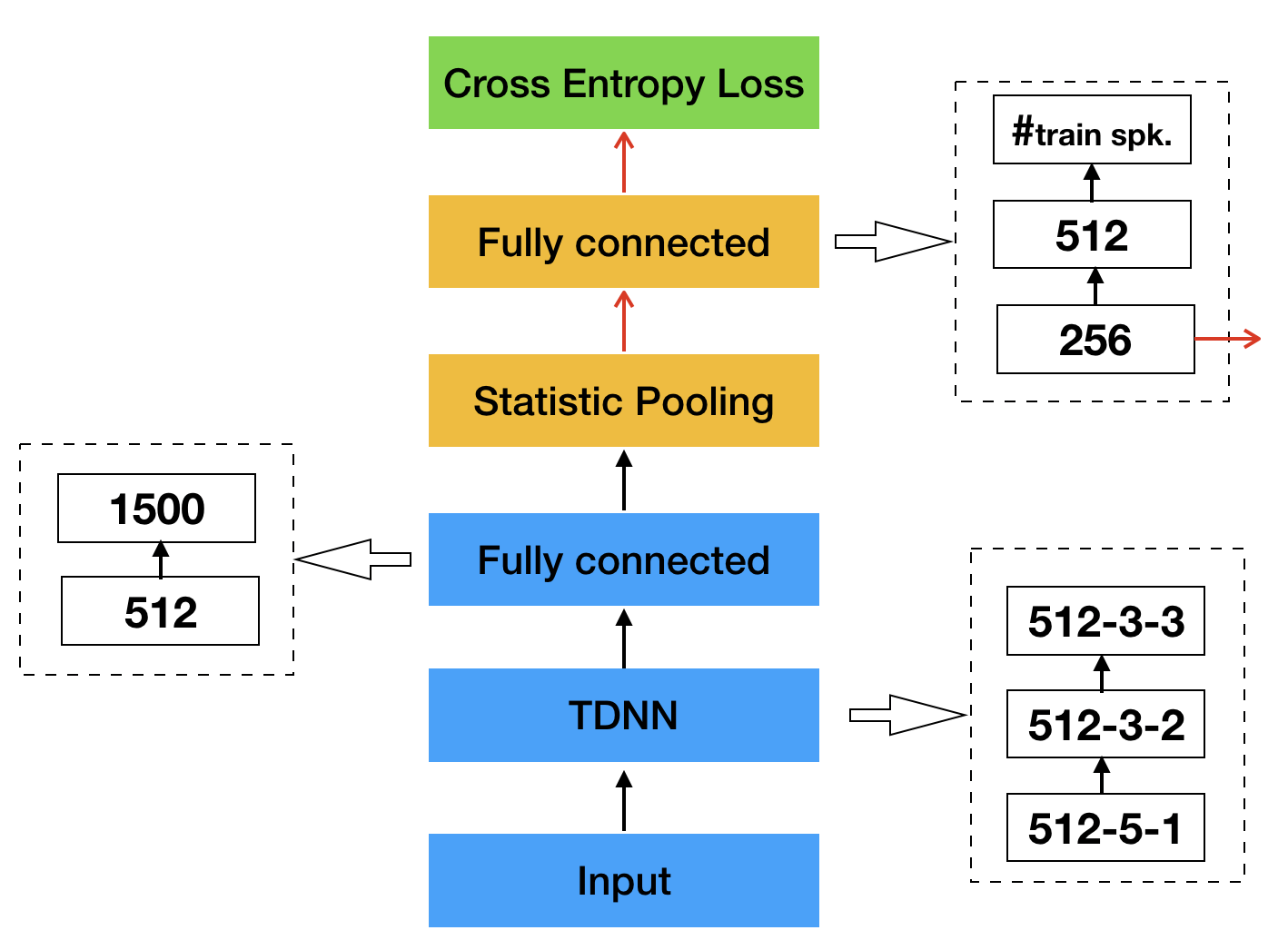}}
\caption{Model structure of baseline x-vector system.}
\label{fig:xvector}
\end{figure}

Figure \ref{fig:xvector} depicts the deep neural network configuration for x-vector system \cite{snyder2017deep}\cite{snyder2018x}. The blue, yellow and green blocks represent the frame level, utterance level and training loss used in the system training. Black arrows indicate a ReLU activation function. Batch normalization layers are added in two adjacent layers, while the red arrows indicate there is no nonlinear mapping added between layers.  

The frame level model consists of three TDNN layers to extract speaker information at the frame level. 
X-Y-Z in the block represents the number of filters, filter window size and dilation number at this layer. 
At the utterance level, high order stats pooling methods such statistical pooling or attentive statistical pooling can be used. The output (concatenation of the mean and standard deviation vectors) is fed into another three fully connected layers. Speaker embeddings are then extracted from the output of the first linear projection layer.   

Cross entropy loss is used to train the system and reduce the confusion between all speakers in the training set. Once speaker embeddings are extracted, LDA and PLDA \cite{ioffe2006probabilistic} are used as back-end scoring, as in the standard Kaldi x-vector recipe.
\subsection{Related work}
Combining CNN or TDNN with LSTM is proved effective in automatic speech recognition (ASR) tasks.  Sainath et al. \cite{sainath2015convolutional} proposed to stack CNN, LSTM and DNN sequentially for speech recognition and it shows superior results than using CNN, LSTM or DNN alone.  Peddinti et al. \cite{peddinti2018low} conducted experiments to compare the stack order of TDNN and LSTM and found interleaving of TDNN layers with LSTM layers could be more effective than simple stacking strategy. 

Pooling module is the key component to bridge frame level features to speaker representation. High order statistical pooling \cite{snyder2017deep} shows better performance than simple pooling method like average or maximum pooling. Okabe et al. \cite{okabe2018attentive} proposed  parametric based attentive pooling to give different importance for each frame feature. Zhu et al. \cite{zhu2018self} further extended it to multiple heads to generate utterance representation from different views. The multiple-level pooling proposed in this work is focused on the way to gather pooling features instead of pooling module itself.    

\section{Deep Speaker Embedding with Multi-level pooling}
\label{sec:sys_des}
\begin{figure}[htb]
  \centering
  \centerline{\includegraphics[width=6.0cm]{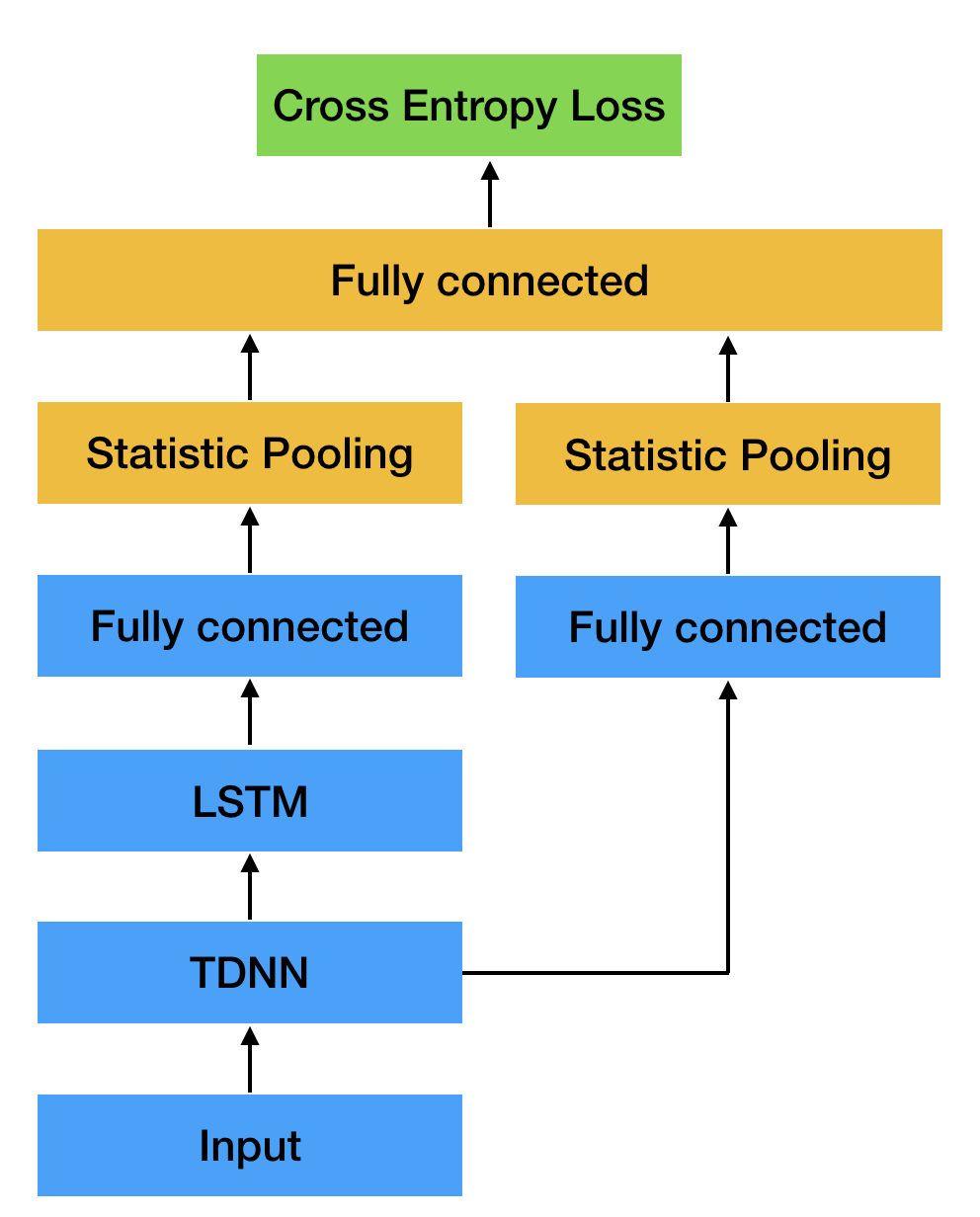}}
\caption{Deep speaker embedding with multi-level pooling (Model $MP$ in Table~\ref{tbl:mdl_cfg})}
\label{fig:mul-stage}
\end{figure}

Figure  \ref{fig:mul-stage} shows the proposed new deep speaker embedding training used in this study. 
In order to generate complimentary information, a hybrid structure with both TDNN and LSTM layers is proposed in this framework. The TDNN layer would focus on the local feature representation, while the LSTM layer would consider global and sequential information from the whole utterance. The multi-level pooling from the TDNN and LSTM layers generate different representations on the utterance level. This would help to gather speaker information from different spaces and generate a comprehensive speaker representation. This would also help to pass error signals back to earlier layers which, in turn, helps alleviate the vanishing gradient problem. The outputs of different pooling layers are concatenated and fused into the following fully connected layers. Even though residual networks with very deep neural network structures \cite{He_2016}\cite{li2018deep} are popular for computer vision tasks, we haven't seen good improvements from very deep structures, as shown in \cite{novoselov2018deep}. Therefore we use the same number of TDNN layers as in x-vector model.




The x-vector system is not an end-to-end SV system. The extracted speaker embeddings are passed to the back-end classifiers of LDA and PLDA, which are both linear transforms trained with different objective functions. Thus there is a mismatch of training loss and LDA/PLDA training objectives, as noticed in \cite{li2018full}. In order to make the embedding output suitable for the backend, a regularization term is introduced by applying a constraint on the norm of the embedding layer output. 
\begin{equation} \label{equ_norm}
\mathcal{L} = -\sum_{i=1}^{M}log\frac{\exp^{{w_{c_i}^T}x_i+b_{c_i}}}{\sum_j^N\exp^{{w_{j}^T}x_i+b_j}} + \lambda ||z_i||_2
\end{equation}
 In the above equation, $c_i$ is the speaker index from training sample $i$, $w_j$ corresponds to j-th column of $W \in \mathcal{R}^{dXN}$ the last linear projection layer before softmax, $b_j$ is the bias term, $N$ and $M$ are the number of speakers and samples in the training data respectively. $x_i$ is the output tensor for the 2nd fully connected layer in the utterance level model, $z_i$ the output tensor of the 1st fully connected layer without ReLU non-linearity and batch normalization. Note, the regularization is applied to the output of the speaker embedding extraction layer directly instead of the weights in the neural network. 
    
 
\section{Experiments and Results}
\label{sec:expt}
\subsection{Data set }\label{ssec:data}
Both the baseline x-vector and our proposed deep speaker embedding systems are evaluated on the NIST SRE16 eval test and SRE18 dev test. The SRE16 test set includes Tagalog and Cantonese telephone speech. For the SRE18 dataset, only Call My Net2 (CMN2) portion of development data set is used in our test. The CMN2 data is composed of both PSTN and VOIP data collected outside North America, spoken in Tunisian Arabic. 
The amount of enrollment data is around 60 seconds per speaker and the duration test data is ranging approximately from 10 seconds to 60 seconds. In both SRE16 and SRE18, the tasks aim to deal with domain mismatch between the development and test data, including channel, noise condition and language mismatches.

In order to handle multiple domain data, a large training dataset is formed from SRE18 allowed, publicly available, data sources: SRE data from years 2004 to 2006, 2008, 2010; all Switchboard data; all Fisher data and all Voxceleb data, with a total amount of $13,564$ hours. This dataset consists of both telephone and microphone speech from total $20,803$ speakers. The model is trained on the 8kHz data. The features include 40 dimensional filterbanks and 3 dimensional pitch features. The feature has frame-lengths of 25ms, mean-normalized over a sliding window of up to 3 seconds. The Kaldi energy based SAD is used to filter out non-speech frames.
  
We follow the same data augmentation procedure conducted in \cite{snyder2018x} on the training data to alleviate the noise condition mismatch. The data augmentation strategy includes both adding additive noises and reverberation data.  

In both SRE16 and SRE18 NIST provided tens of hours unlabelled but matched data for development.  
For back-end scoring, an LDA projects speaker embeddings from 256 to 150, estimated from SRE training data. The same SRE training data is also used to estimate the PLDA classifier. An extra PLDA adaptation \cite{alam2018speaker} is conducted on the unlabelled development data to reduce training/testing mismatch.  


\subsection{Experimental Setup}\label{ssec:setup}
The configuration of the baseline x-vector system is described in Section \ref{ssec:baseline}. 
One thing to mention is that a smaller neuron number is used in the speaker embedding extraction layer, say 256 instead of 512, as shown in Figure \ref{fig:xvector}. This benefits the following backend processing step such as LDA and PLDA estimation to give better evaluation results.
 
The configurations of the three systems to be examined as well as the baseline system are compared in Table \ref{tbl:mdl_cfg}:

\begin{table}[htb]
    \caption{Experimental model configures}\label{tbl:mdl_cfg}
    \bigskip 
    \begin{center}
    \begin{tabular}{l|c}
        \hline
        Model Name & Model Configurations \\
        \hline
        x-vector & TDNN1-TDNN2-TDNN3-P \\
        \hline
        A & TDNN1-P-TDNN2-P-TDNN3-P \\
        \hline
        B & TDNN1-TDNN2-TDNN3-LSTM-P \\
        \hline
        MP & TDNN1-TDNN2-TDNN3-P-LSTM-P \\
        \hline
    \end{tabular}
    \end{center}
\end{table}
In the above table, TDNN$i$ represents the $ith$ TDNN layer in the frame level model, $P$ represents a pooling operation applied to the output of the previous layer. Each pooling operation includes two fully connected layers with statistical pooling afterwards. 
Model $MP$ (Fig. 2) is the proposed system and x-vector is the baseline system in this study. Models $A$ and $B$ are model configurations between  $MP$ and the x-vector baseline. These two models are used to understand the differences between regular pooling and our multi-level pooling. 

Compared with x-vector and model $A$, model $B$ and $MP$ each have one bidirectional LSTM layer after TDNN blocks in the frame level model. The number of neurons is 512. x-vector and model $B$ are single pooling based systems while model $A$ and $MP$ are multiple pooling based systems. In the multiple pooling based systems ($A$ and $MP$), different statistical pooling outputs will be concatenated together before sending to the following fully connected layers in the utterance level model. 
In order to keep the size of pooling output across different systems the same,  the number of neurons in the 2nd fully connected layer (before pooling module) will be changed from 1500 to 500 and 750 for model $B$ and $MP$ respectively. 

The system is implemented with PyTorch and optimized with stochastic gradient descent (SGD) using momentum (0.9), with a mini-batch size of 256.

\subsection{Experimental Results}\label{ssec:results}
We report results in terms of EER and the minimum DCF at $P_{target}=0.01$ and $P_{target}=0.005$. No score normalization, e.g., t-norm or s-norm \cite{dehak2010cosine}, is applied in this study.

\begin{table}[!htb]
	\caption{Results on SRE16 eval test.}
	\label{tbl:sre16}
	\bigskip
	\begin{center}
		\vskip -0.1in
		\setlength{\tabcolsep}{2pt}
    \begin{tabular}{l|l|l|l||r|r|r}
    \hline
    Model &
      \multicolumn{3}{c||}{$\lambda=0$} &
      \multicolumn{3}{c}{$\lambda=0.001$} \\
      
    & Pooled & Tag. & Can. & Pooled & Tag. & Can. \\
    \hline
        x-vector &    7.61 & 10.98 &  3.95 & 6.90 & 10.20 & 3.50 \\
        \hline
        A &    8.17 & 11.78 &  4.45 & 7.51 & 10.93 & 4.06 \\
        \hline
        B & \textbf{6.64} & 9.80 & \textbf{3.40} & 6.84 & 9.83 & 3.82\\
        \hline
        MP &   6.68  &  \textbf{9.74} &  3.51  & \textbf{6.13} & \textbf{9.18} &  \textbf{3.13} \\
        \hline
    \end{tabular}
    \end{center}
\end{table}
The EERs on the SRE16 test set are shown in Table \ref{tbl:sre16}. The left side of the table are results without regularization and the right side with regularization $\lambda=0.001$.

We first examine our results when no regularization is applied. Our results from the left side of Table 1 shows that with no regularization, there is no gain with pooling comparing Model $A$ to the baseline x-vector. Adding a bidirectional LSTM in Model $B$ helps reducing the EER by 12.7\% relative to the baseline.

Now we move on to our results with regularization (the right side of Table 2). Compared to the results on the left side, most models achieve significant EER reduction. Our proposed model ($MP$) achieves the best results in this test set. These results show that multiple pooling operations is only beneficial if the pooling information comes from different sources. In our proposed model $MP$, one pooling operation is conducted after the TDNN layers, which focus on local information, and the 2nd pooling operation is conducted after the LSTM layer, which extracts sequential information.

Adding a regularization term on the norm of the embedding output is useful for most models, especially for models with multiple pooling. This reduces the range of the norm of the output embedding vectors, from 500 to 10. This makes the speaker embeddings more numerically stable for the backend scoring. Reducing the norm of the output embeddings also may condense features from different levels of the model, into the same numerical range, which helps them to be fused properly. 

Overall, comparing the result from our proposed model $MP$ with regularization, against the baseline x-vector model, we observe a relative 19\% EER reduction. The corresponding DCF scores are reported in Table \ref{tbl:sre16_dcf}, for different $P_{target}$ values. There is a 14.6\% and 12.3\% DCF score reduction compared with the baseline for $P_{target}$ = 0.01 and 0.005 respectively.

\begin{table}[t]
    \caption{DCF scores for SRE16 (pooled) test set}\label{tbl:sre16_dcf}
    \bigskip
    \begin{center}
    \begin{tabular}{l|c|c|c|c}
        \hline
        \multirow{2}{*}{model}  & \multicolumn{2}{c|}{ $\lambda=0$ } & \multicolumn{2}{c}{ $\lambda=0.001$} \\
        \cline{2-5}
        &  $p=0.01$ & $p=0.005$ & $p=0.01$ & $p=0.005$ \\
        \hline
        x-vector &  0.593  &  0.651 & 0.594 &  0.673 \\
        \hline
        B      & 0.581   &  0.656 & 0.525 &  0.586  \\
        \hline
        MP     & 0.567   &  0.632 & \textbf{0.506} &   \textbf{0.571}  \\
        \hline
    \end{tabular}
    \end{center}
\end{table}

\begin{table}[t]
    \caption{Evaluation results on the SRE18 (CMN2) dev set}\label{tbl:sre18}
    \bigskip 
    \begin{center}
    \begin{tabular}{l|l|c|c|c}
        \hline
         $\lambda$ & model & EER & $DCF(0.01)$ & $DCF(0.005)$ \\ 
        \hline
        \multirow{ 2}{*}{$0.0$} &x-vector & 7.29 & 0.593  & 0.651  \\
        \cline{2-5}
        & B & 7.90 & 0.581 &  0.656  \\
        \cline{2-5}
        & MP&  7.16 & 0.567 &  0.632   \\
        \hline
        \multirow{3}{*}{$0.001$} & x-vector & 7.46 & 0.594 &  0.673 \\
        \cline{2-5}
        & B & 7.77 & 0.525 &  0.586 \\
        \cline{2-5}
        & MP&  \textbf{6.61} & \textbf{0.506} &   \textbf{0.571}    \\
        \hline
    \end{tabular}
    \end{center}
\end{table}

In Table \ref{tbl:sre18}, results on the SRE18 CMN2 development set are presented. Similar to SRE16 results, model $MP$ achieves the best result: 9.3\% EER reduction and 12\% to 14\% DCF score reduction are achieved when regularization is applied.

\section{Conclusions}
In this study, we propose a novel pooling strategy for gathering speaker information from different levels of the model. Our three main contributions are as follows: 
(1) A hybrid model structure including both TDNN and LSTM is employed to generate complimentary information. Output from TDNN blocks focuses on local information and LSTM layer emphasizes on global and sequential information generated from the whole utterance. 
(2) Different representation extract from the hybrid model are pooled at multiple levels and combined to obtain a robust speaker representation.
(3) A regularization is applied to the output of the embedding extraction layer, which significantly reduces the norm of extracted embedding and keeps it in a reasonable value range as well as keeping the embedding more numerically stable for the following backend scoring. 
The synergy of these improvements are shown on the NIST SRE 2016 eval test (with a 19\% EER reduction) and SRE 2018 dev test (with a 9\% EER reduction), as well as more than 10\% DCF scores reduction on these two test sets over the x-vector baseline. 
\label{sec:con_dis}

\vfill\pagebreak

\bibliographystyle{IEEEbib}
\bibliography{refs}

\begin{thebibliography}{10}

\bibitem{hansen2015speaker}
John~HL Hansen and Taufiq Hasan,
\newblock ``Speaker recognition by machines and humans: A tutorial review,''
\newblock {\em IEEE Signal processing magazine}, vol. 32, no. 6, pp. 74--99,
  2015.

\bibitem{heigold2016end}
Georg Heigold, Ignacio Moreno, Samy Bengio, and Noam Shazeer,
\newblock ``End-to-end text-dependent speaker verification,''
\newblock in {\em Acoustics, Speech and Signal Processing (ICASSP), 2016 IEEE
  International Conference on}. IEEE, 2016, pp. 5115--5119.

\bibitem{snyder2017deep}
David Snyder, Daniel Garcia-Romero, Daniel Povey, and Sanjeev Khudanpur,
\newblock ``Deep neural network embeddings for text-independent speaker
  verification,''
\newblock in {\em Proc. Interspeech}, 2017, pp. 999--1003.

\bibitem{snyder2018x}
David Snyder, Daniel Garcia-Romero, Gregory Sell, Daniel Povey, and Sanjeev
  Khudanpur,
\newblock ``X-vectors: Robust dnn embeddings for speaker recognition,''
\newblock {\em ICASSP}, 2018.

\bibitem{Dehak2011FrontEndFA}
Najim Dehak, Patrick Kenny, Reda Dehak, Pierre Dumouchel, and Pierre Ouellet,
\newblock ``Front-end factor analysis for speaker verification,''
\newblock {\em IEEE Transactions on Audio, Speech, and Language Processing},
  vol. 19, pp. 788--798, 2011.

\bibitem{li2018deep}
Na~Li, Deyi Tuo, Dan Su, Zhifeng Li, and Dong Yu,
\newblock ``Deep discriminative embeddings for duration robust speaker
  verification,''
\newblock {\em Proc. Interspeech 2018}, pp. 2262--2266, 2018.

\bibitem{novoselov2018deep}
Sergey Novoselov, Andrey Shulipa, Ivan Kremnev, Alexandr Kozlov, and Vadim
  Shchemelinin,
\newblock ``On deep speaker embeddings for text-independent speaker
  recognition,''
\newblock in {\em Odyssey}, 2018.

\bibitem{okabe2018attentive}
Koji Okabe, Takafumi Koshinaka, and Koichi Shinoda,
\newblock ``Attentive statistics pooling for deep speaker embedding,''
\newblock {\em arXiv preprint arXiv:1803.10963}, 2018.

\bibitem{zhu2018self}
Yingke Zhu, Tom Ko, David Snyder, Brian Mak, and Daniel Povey,
\newblock ``Self-attentive speaker embeddings for text-independent speaker
  verification,''
\newblock {\em Proc. Interspeech 2018}, pp. 3573--3577, 2018.

\bibitem{li2017deep}
Chao Li, Xiaokong Ma, Bing Jiang, Xiangang Li, Xuewei Zhang, Xiao Liu, Ying
  Cao, Ajay Kannan, and Zhenyao Zhu,
\newblock ``Deep speaker: an end-to-end neural speaker embedding system,''
\newblock {\em arXiv preprint arXiv:1705.02304}, 2017.

\bibitem{Wan2018GeneralizedEL}
Li~Wan, Quan Wang, Alan Papir, and Ignacio Lopez-Moreno,
\newblock ``Generalized end-to-end loss for speaker verification,''
\newblock {\em 2018 IEEE International Conference on Acoustics, Speech and
  Signal Processing (ICASSP)}, pp. 4879--4883, 2018.

\bibitem{novoselov2018triplet}
Sergey Novoselov, Vadim Shchemelinin, Andrey Shulipa, Alexandr Kozlov, and Ivan
  Kremnev,
\newblock ``Triplet loss based cosine similarity metric learning for
  text-independent speaker recognition,''
\newblock {\em Proc. Interspeech 2018}, pp. 2242--2246, 2018.

\bibitem{huang2018angular}
Zili Huang, Shuai Wang, and Kai Yu,
\newblock ``Angular softmax for short-duration text-independent speaker
  verification,''
\newblock {\em Proc. Interspeech 2018}, pp. 3623--3627, 2018.

\bibitem{ioffe2006probabilistic}
Sergey Ioffe,
\newblock ``Probabilistic linear discriminant analysis,''
\newblock in {\em European Conference on Computer Vision}. Springer, 2006, pp.
  531--542.

\bibitem{sainath2015convolutional}
Tara~N Sainath, Oriol Vinyals, Andrew Senior, and Ha{\c{s}}im Sak,
\newblock ``Convolutional, long short-term memory, fully connected deep neural
  networks,''
\newblock in {\em Acoustics, Speech and Signal Processing (ICASSP), 2015 IEEE
  International Conference on}. IEEE, 2015, pp. 4580--4584.

\bibitem{peddinti2018low}
Vijayaditya Peddinti, Yiming Wang, Daniel Povey, and Sanjeev Khudanpur,
\newblock ``Low latency acoustic modeling using temporal convolution and
  lstms,''
\newblock {\em IEEE Signal Processing Letters}, vol. 25, no. 3, pp. 373--377,
  2018.

\bibitem{He_2016}
Kaiming He, Xiangyu Zhang, Shaoqing Ren, and Jian Sun,
\newblock ``Deep residual learning for image recognition,''
\newblock in {\em 2016 IEEE Conference on Computer Vision and Pattern
  Recognition (CVPR)}, Jun 2016.

\bibitem{li2018full}
Lantian Li, Zhiyuan Tang, Dong Wang, and Thomas~Fang Zheng,
\newblock ``Full-info training for deep speaker feature learning,''
\newblock in {\em 2018 IEEE International Conference on Acoustics, Speech and
  Signal Processing (ICASSP)}. IEEE, 2018, pp. 5369--5373.

\bibitem{alam2018speaker}
Md~Jahangir Alam, Gautam Bhattacharya, and Patrick Kenny,
\newblock ``Speaker verification in mismatched conditions with frustratingly
  easy domain adaptation,''
\newblock in {\em Proc. Odyssey 2018 The Speaker and Language Recognition
  Workshop}, 2018, pp. 176--180.

\bibitem{dehak2010cosine}
Najim Dehak, Reda Dehak, James~R Glass, Douglas~A Reynolds, and Patrick Kenny,
\newblock ``Cosine similarity scoring without score normalization
  techniques.,''
\newblock in {\em Odyssey}, 2010, p.~15.

\end{thebibliography}

\end{document}